\documentclass[sigconf,nonacm]{acmart}

\usepackage{amsfonts}
\usepackage{nicefrac}
\usepackage{multirow}
\usepackage{array}
\usepackage{placeins}
\usepackage{algorithm}
\usepackage{algorithmic}
\renewcommand\footnotetextcopyrightpermission[1]{}
\setcopyright{none}
\acmConference[arXiv]{arXiv preprint}{2026}{}
\acmYear{2026}
\acmDOI{}
\acmISBN{}

\title[Self-Evolution via Divergence-Point Preferences]{Self-Evolution for Multi-Turn Tool-Calling Agents via Divergence-Point Preference Learning}

\author{Jiaqiang Tang}
\affiliation{%
  \institution{The Hong Kong University of Science and Technology (Guangzhou)}%
  \city{}%
  \country{}%
}
\email{jtang383@connect.hkust-gz.edu.cn}


\begin{document}

\begin{abstract}
Multi-turn tool-using agents must coordinate long-horizon tool sequences while tracking dialogue state and policy constraints. Existing approaches often separate inference-time orchestration from parameter-level learning, leaving tool selection weakly structured and preference updates vulnerable to train--deployment prompt mismatch. For within-benchmark self-improvement, ToolGraph combines schema-derived topology, transition weights estimated from successful rollouts, and history-aware controls for write prerequisites and repeated-search loops. We then construct 161 preference pairs by locating divergence points via state-based matching and prefix-based alignment, filtered through action-correctness annotations, and train DPO under the same ToolGraph context used at inference. Across 375 tau2-bench tasks, ToolGraph raises the weighted average reward from 0.304 to 0.338 (+11.2\% relative), while ToolGraph+DPO reaches 0.355 (+16.8\% over the baseline), with the DPO gain concentrated in airline and retail. Fine-grained diagnostics further show that roughly half of telecom trajectories exhaust the step budget before action execution and that chosen reward positivity is the most useful checkpoint signal across our 16 evaluated DPO configurations.
\end{abstract}

\maketitle

\begin{figure*}[!t]
\centering
\includegraphics[width=0.90\textwidth]{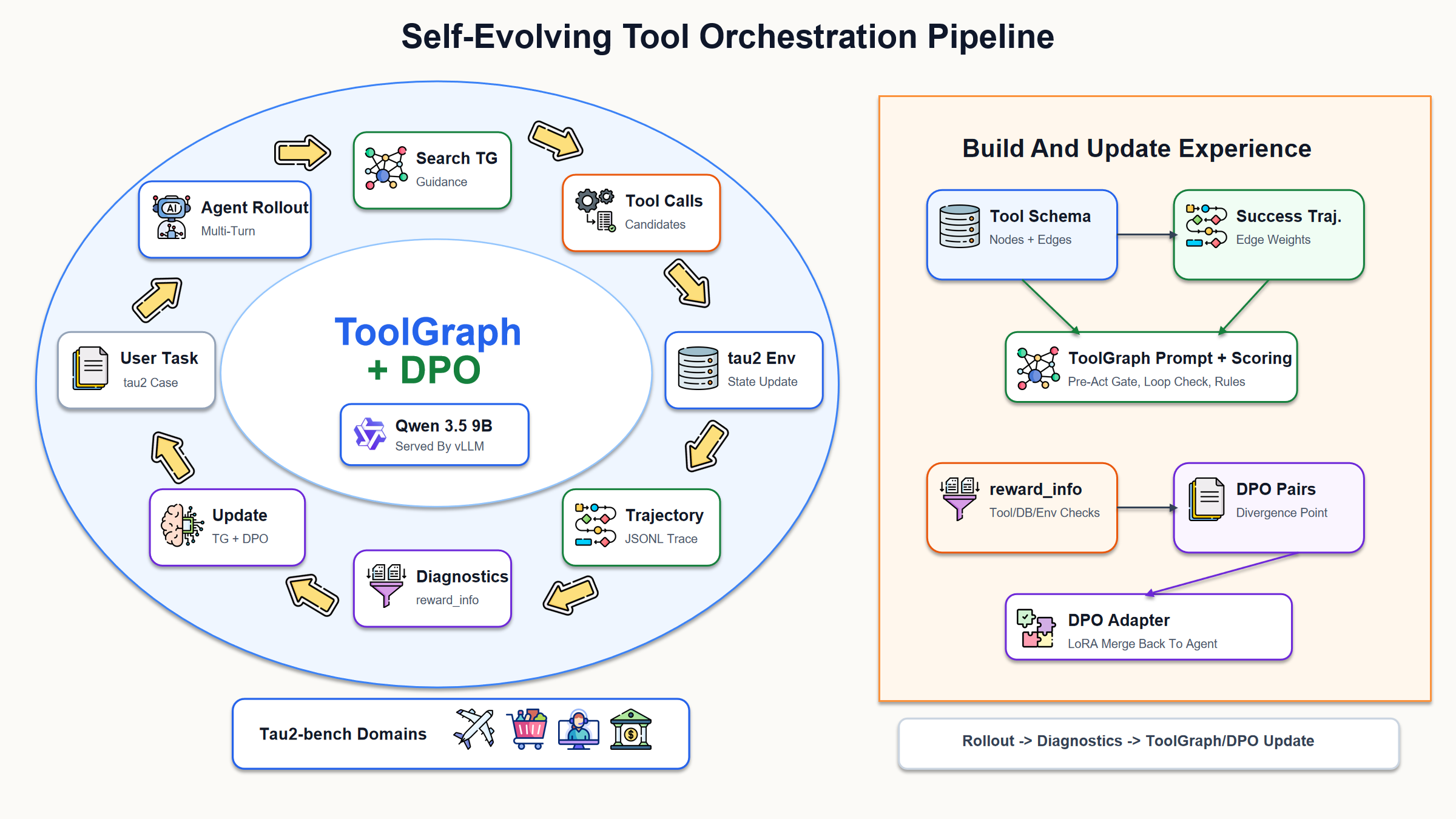}
\Description{Pipeline diagram showing a vLLM-served Qwen agent, ToolGraph-guided tool selection, trajectory logging, failure diagnosis, DPO pair construction, and adapter update.}
\caption{System pipeline overview. The agent runs on vLLM-served Qwen 3.5 9B (execution layer). ToolGraph and the DPO pipeline form the control layer, providing structural guidance and parameter-level self-evolution. The instrumentation layer records trajectories and intermediate reward signals for analysis and training.}
\label{fig:system-overview}
\end{figure*}


\section{Introduction}

Multi-turn customer-service agents must use domain-specific tool sets, track conversation state, follow policy constraints, and decide when to stop gathering information and execute an action. The tau2-bench benchmark~\citep{tau2bench2025} captures this challenge through four domains: airline (50 tasks), retail (114), telecom (114), and banking (97). An assistant communicates with a simulated user and invokes domain tools while the environment evaluates the final state. Deploying a local model (Qwen/Qwen3.5-9B~\citep{qwen35_9b_hf}) on this benchmark requires reliable tool selection across a large API surface and a self-evolution loop that turns execution experience into better policy.

This problem is challenging for two reasons. First, the tool surface is large and unstructured: each domain exposes 13--16 tools with overlapping parameters, and the agent must choose correct sequences across multiple turns without explicit supervision. Second, the agent must learn from its own execution traces without turning incidental trajectory differences into incorrect preference signals. Parameter-level updates are especially fragile when the training context differs from deployment: our initial DPO setting degraded online reward when the adapter was trained without ToolGraph's planning prompt but deployed with it~\citep{rafailov2023dpo}.

In this paper, we investigate the following question: \emph{Can an agent reuse its own benchmark experience as both inference-time structure and prompt-aligned preference supervision?} Our system addresses this question through two coordinated components: (1) ToolGraph orchestration, whose graph structure comes from tool schemas and whose edge weights come from successful trajectories, and (2) DPO self-evolution that converts the agent's successful and failed trajectories into preference updates under the same ToolGraph context used at inference. Figure~\ref{fig:system-overview} illustrates the pipeline.

\textbf{ToolGraph orchestration.} ToolGraph represents each domain's tools as a weighted graph. The topology is built from tool schemas and shared parameters, while edge probabilities are computed from the agent's own successful trajectories. On top of this graph, we add three structural mechanisms: a pre-act gate for write-prerequisite rules, a telecom loop check for repeated line searches, and Critical Rules distilled from successful trajectory patterns. Unlike static schema-only guidance, ToolGraph combines schema structure with experience-derived edge weights.

Structural guidance alone does not cover all boundary cases and policy subtleties. ToolGraph raises the weighted average from 0.304 to 0.338, but gains plateau: retail remains below 0.30, airline below 0.67, and much of the improvement comes from suppressing MAX\_STEPS terminations rather than learning better action sequences. This motivates parameter-level learning from experience.

\textbf{DPO self-evolution under ToolGraph.} Preference pairs are constructed in the same ToolGraph context where the model is deployed. This aligns the training and deployment contexts when combining TG and DPO. For each task with both a success and a failure trajectory, the framework locates the divergence point---the turn where the two trajectories take different actions---through state-based matching (Stage~1) or prefix-based alignment (Stage~2). The success action becomes \textit{chosen} and the failure action becomes \textit{rejected}, after passing through an annotation-based correctness filter that screens pairs against tau2-bench's action-correctness signals.

Across the evaluated 16 DPO configurations, \emph{chosen reward positivity} is the most useful health indicator we observe: configurations where chosen reward remains positive improve online reward, while configurations with negative chosen reward show catastrophic forgetting despite higher preference margins.

We evaluate on all 375 tau2-bench tasks using Qwen/Qwen3.5-9B served by vLLM on HPC2 A800 GPUs. ToolGraph raises the weighted average reward from 0.304 to 0.338, and DPO R2\#4 further improves it to 0.355. Relative to ToolGraph, DPO improves task-level mean reward on 12 airline tasks and 28 retail tasks while regressing on 7 and 18, respectively. Telecom's dominant observed bottleneck, revealed through tau2-bench's intermediate diagnostics, is a \texttt{MAX\_STEPS} loop: about half of the trajectories terminate before action execution, while action accuracy reaches 95\% in the diagnostic subset that reaches the action phase.

Our main contributions are three-fold. \textbf{(1)} We develop ToolGraph, an orchestration layer that converts tool schemas and prior successful trajectories into a weighted dependency graph with history-aware beam scoring, domain-specific write-prerequisite controls, and loop detection. \textbf{(2)} We design a preference-pair construction framework that extracts chosen--rejected action pairs at trajectory divergence points, trains DPO under the same ToolGraph context used at deployment, and filters pairs via benchmark action-correctness annotations to isolate causal preference signals. \textbf{(3)} We identify two diagnostic findings: telecom's bottleneck lies in pre-action step-budget exhaustion---not action accuracy---with nearly half of trajectories terminating before reaching the action phase; and chosen reward positivity is a more reliable checkpoint-selection signal than accuracy or margin across the evaluated DPO configurations.


\section{Related Work}

\textbf{Tool-using conversational agents.}
ReAct-style agents interleave reasoning and tool actions to solve interactive tasks~\citep{yao2023react}. tau2-bench evaluates this capability in a dual-control setting where a simulated user and environment expose whether the agent can gather information, follow policy, and execute tools correctly~\citep{tau2bench2025}. Recent work on tau-bench-style tasks studies prompt reformulation and checklist-style memory as ways to improve tool-use accuracy~\citep{mishra2025irma}. Our work instead focuses on structured orchestration: ToolGraph converts tool schemas and successful trajectories into graph guidance, then uses best-of-N scoring to bias tool choices during inference. This is related in spirit to search-based reasoning methods such as Tree of Thoughts~\citep{yao2023tree}, but our search space is tool-call sequences in a live benchmark rather than abstract reasoning states. Beyond text-based tasks, self-evolution has been applied to computer-use agents: EvoCUA evolves GUI agents through iterative synthetic task generation and sandbox rollouts, showing that experience-driven improvement can transfer across domains~\citep{xue2025evocua}.

\textbf{Preference learning and self-evolution.}
Self-evolving agents aim to improve from their own interaction experience rather than relying only on static supervised data~\citep{fang2025selfevolving}. SPIN iteratively trains a model to distinguish its own responses from human demonstrations through self-play, without additional annotation~\citep{chen2024spin}. Reflexion uses verbal feedback as a form of episodic memory for language agents~\citep{shinn2023reflexion}. ETO converts successful and failed agent trajectories into trajectory-level DPO pairs~\citep{song2024eto}, while EvolveR distills historical trajectories into reusable experience principles and trains agents to use them through an experience-driven lifecycle~\citep{wu2025evolver}. Our system follows the same broad direction, but stores experience in structured ToolGraph parameters and constructs local divergence-point preferences rather than whole-trajectory preferences. Direct Preference Optimization (DPO) turns pairwise preferences into a supervised objective without an explicit reward model~\citep{rafailov2023dpo}, and QLoRA makes such adaptation feasible under limited GPU memory~\citep{dettmers2023qlora}. This paper studies an engineering requirement for tool agents: preference data should be screened against available action-level correctness annotations and trained under the same ToolGraph context used at deployment.


\section{Problem Setup}
\label{sec:problem-setup}

We formulate a tau2-bench interaction as a partially observable Markov decision process (POMDP)
\[
\mathcal{M}=(\mathcal{S},\mathcal{A},\mathcal{T},\mathcal{R},\Omega,\mathcal{O}).
\]
The latent state $s_t\in\mathcal{S}$ contains the simulated user's goal, the domain environment and database state, and the interaction state at turn $t$. An agent action $a_t\in\mathcal{A}$ is either a natural-language response or a set of one or more tool calls with arguments. The transition function $\mathcal{T}(s_{t+1}\mid s_t,a_t)$ captures the joint response of the user simulator, domain tools, and environment to the agent's action. The benchmark reward function $\mathcal{R}(s_t,a_t)$ evaluates task completion; in our runs, the reported trajectory reward is terminal, so $R(\tau)=r_T$. The observation space $\Omega$ contains user messages and tool results, and $\mathcal{O}(o_t\mid s_t)$ denotes the observation function that exposes $o_t\in\Omega$ without revealing the complete simulator state.

Because the environment state is partially observed, the policy conditions on the interaction history $h_t=(o_0,a_0,\ldots,o_t)$. ToolGraph additionally supplies a domain-specific planning context $c_d$, giving the deployed policy $\pi_\theta(a_t\mid h_t,c_d)$. We keep $c_d$ separate from the environment observation: it is control-layer guidance constructed from tool schemas and prior trajectories, not privileged access to $s_t$. A trajectory is
\[
\tau=(o_0,a_0,o_1,a_1,\ldots,o_T),
\qquad
J(\pi_\theta)=\mathbb{E}_{\tau\sim\pi_\theta}[R(\tau)].
\]
The benchmark's intermediate \texttt{reward\_info} fields are used for diagnosis and annotation-based correctness screening, not as additional online rewards. Given a collection of prior trajectories $\mathcal{D}$ from the same benchmark task distribution, our self-evolution objective is to increase $J$: ToolGraph estimates external tool-transition guidance from $\mathcal{D}$, while DPO updates $\pi_\theta$ from chosen--rejected action pairs. This is a within-benchmark self-improvement setting; it does not measure generalization to unseen tasks or domains.


\section{Method}
\label{sec:method}

\subsection{Overall Architecture}
\label{sec:architecture}

The system wraps tau2-bench with a local OpenAI-compatible agent runner and is organized into three layers. The \textbf{execution layer} handles model serving and response generation. The 9B Qwen model runs on HPC2 A800 GPUs through vLLM~\citep{vllm2023} under SLURM. The \textbf{control layer} selects interventions: ToolGraph planning prompts injected into the system prompt, best-of-N tool-call selection with graph-derived scoring (Section~\ref{sec:toolgraph}), and DPO adapter weights applied at inference time through LoRA merging (Section~\ref{sec:dpo}). The \textbf{instrumentation layer} records per-step trajectories, tool-call details, termination reasons, and tau2-bench's fine-grained \texttt{reward\_info} object (action-level match flags, DB match status, environment assertion results). All methods share the same logging format, enabling controlled comparisons.

The architecture evolved from v1 to v2 primarily by making TG and DPO consistent: v1 trained DPO with a bare system prompt but inferred with TG-injected prompts, which made the learned preferences inconsistent with deployment. In v2, DPO data is constructed under the same ToolGraph context used at inference. On the ToolGraph side, v1's schema-only graph with zero transition weights was extended with edge weights from 532 curated success trajectories and domain-specific scoring mechanisms (pre-act gate, loop detection, backtracking rewards). The graph topology and weights are data-derived, whereas Critical Rules, phase classification, and scoring adjustments are manually specified heuristics based on observed task patterns; the resulting controller is therefore not a fully automatic discovery system.

\subsection{ToolGraph Orchestration}
\label{sec:toolgraph}

ToolGraph is a domain-specific directed dependency graph that provides structural guidance for multi-turn tool selection. Its construction includes schema-derived edges, weights estimated from the agent's successful trajectories, a planning prompt, and a beam-scoring mechanism that re-ranks candidate tool calls at inference time.

\subsubsection{Graph Construction and Edge Weights}

ToolGraph represents each domain tool as a node carrying name, description, parameters, and required-argument set. Edges are built through shared-parameter analysis: for every parameter name appearing in two or more tools, directed edges are created between all tool pairs sharing that parameter. Each tool is classified into one of three execution-phase categories via keyword matching: \textbf{identify} tools for user or entity lookup, \textbf{act} tools for write operations such as update, cancel, or booking, and \textbf{gather} tools for remaining information lookups. \textbf{Backward edges} from act tools to gather tools are added when they share an entity parameter (e.g., \texttt{reservation\_id}), enabling the agent to return to information-gathering after a write attempt.

A key limitation of schema-only edges is that they encode structural adjacency but not empirical information about which transitions are useful. We compute edge weights from 532 reward-1.0 trajectories from Qwen 3.5 9B's best runs (92 airline, 226 retail, 192 telecom, 22 banking). Two statistics are combined: \textbf{tool success rate} (proportion of trajectories containing a tool that eventually succeed) and \textbf{ToolGraph transition probability} $P_{\mathrm{TG}}(\text{tgt} \mid \text{src})$ (how often tool \textit{tgt} immediately follows \textit{src} in success trajectories). A \textbf{per-task downweighting} scheme ensures each task contributes equal total weight regardless of rollout count. The final edge score is $\text{score} = P_{\mathrm{TG}}(\text{tgt} \mid \text{src}) \times \max(\text{success\_rate}(\text{tgt}), 0.01)$. The notation distinguishes empirical tool transitions from the POMDP environment transition $\mathcal{T}$. Cross-validation shows reliable coverage for retail (recall 90\%, precision 64\%) but insufficient coverage for banking (recall 12\%, precision 90\%). For airline and retail, estimated edge probabilities closely track observed transitions ($r=0.91/0.97$; MAE $=0.062/0.048$).

\subsubsection{Planning Prompt Generation}

The planning prompt injects ToolGraph's structural knowledge into the system prompt with three components. \textbf{Execution phases} group tools into identify $\rightarrow$ gather $\rightarrow$ act with guidance hints (``Resolve user identity first'', ``Independent queries can be parallel'', ``Take action based on gathered information''). \textbf{Parallel sets} identify groups of read-only tools with non-overlapping parameters that can be called simultaneously. \textbf{Critical Rules} encode manually specified domain controls informed by observed successful and failed trajectory patterns: for airline, ``Before cancel/modify: call \texttt{get\_reservation\_details} first''; for telecom, ``After 3+ different line queries: STOP expanding, pick one and fix it.'' The prompt closes with ``Always follow the policy over this guide'' to prevent ToolGraph hints from overriding domain policy.

\subsubsection{Beam Scoring and Best-of-N Selection}
\label{sec:scoring}

While the planning prompt provides soft structural guidance, it cannot prevent suboptimal tool calls within a single turn. We implement best-of-N selection that generates up to two additional candidates via LLM sampling with temperature annealing ($T_i = T_{\text{base}} + i \times 0.15$), deduplicating identical sequences. Each candidate starts at score 1.0 (text-only: 0.5), and five adjustments are applied:

\begin{enumerate}
    \item \textbf{Duplicate penalty} ($-0.25$ per occurrence, capped at $-0.75$ per tool): penalizes tools already called in this conversation.
    \item \textbf{Edge bonus} ($+0.15$): awarded when the first tool call is among top-8 recommended successors of the previous tool.
    \item \textbf{Pre-act gate} ($-0.3$): domain-specific whitelist enforcing write-prerequisite ordering (e.g., \texttt{cancel\_reservation} requires \texttt{get\_reservation\_details} as predecessor).
    \item \textbf{Backtracking reward/penalty} ($\pm 0.10$): a bonus in telecom for re-querying after execution errors, and a penalty in airline and retail for backtracking after act tools.
    \item \textbf{Telecom loop detection} ($-0.3$): penalizes repeated line-detail lookups after three or more distinct line queries.
\end{enumerate}

The highest-scoring candidate is selected; ties favor the candidate with fewer tool calls. This five-layer scoring architecture is the key advance over v1, where zero transition weights rendered the edge bonus ineffective.

\subsection{DPO Self-Evolution}
\label{sec:dpo}

The self-evolution pipeline implements a complete closed loop: the agent runs under ToolGraph guidance, collects trajectories, identifies failures, matches failure states against success trajectories, constructs preference pairs, updates parameters through DPO, and is redeployed. The \textit{chosen} demonstrations in every preference pair are drawn from the agent's \textit{own} success trajectories---not from a stronger model---making this a genuine self-evolution process.

\textbf{Failure attribution.} The module scans all ToolGraph-guided trajectories and classifies low-reward cases into five symptom types: redundant read (identical data re-retrieved), progress stall (three+ consecutive read-only turns past 30\% of steps), tool execution error (environment returned error), max-steps exceeded (50-step budget exhausted), and strategy error (reward $\le 0.5$ with no surface-level symptom). Each case records the symptom type, step index, domain, and task ID; at most two cases per task are retained, prioritized by severity.

\textbf{Structured reflection and pair construction.} Every preference pair is built around a divergence point---the turn where the success and failure trajectories take different actions. The framework locates this point through two strategies, tried in order. \textbf{State-based matching} searches success trajectories for a state compatible with the failure state, using a semantic state signature. \textbf{Prefix-based alignment} falls back to aligning the two trajectories message-by-message. Both strategies produce the same output: a shared prefix serving as prompt $x$, and the diverging actions as $(y^+,y^-)$.

\textbf{State-based divergence.} The framework extracts a state signature from the failure trajectory at the decision point: entity ID, known fields, last tool name, last action type, and pending goal type. It then searches success trajectories of the same task for a compatible state---one sharing the same entity ID, pending goal type, and $\ge 40\%$ Jaccard overlap on known fields. If found and the actions differ, the divergence point is located at the matched state: the shared context up to that point forms the prompt and the diverging actions form the pair. Algorithm~\ref{alg:pair-construction} formalizes the complete procedure. When no compatible state with a different action is found, the algorithm falls through to prefix-based alignment. Shared notation: $\operatorname{System}(d)$ and $\operatorname{TG}(d)$ supply the structural prompt context, $\operatorname{Tok}$ counts tokens, and $\varnothing$ indicates discard.

\begin{algorithm}[H]
\caption{Preference pair construction}
\label{alg:pair-construction}
\footnotesize
\begin{algorithmic}[1]
\STATE \textbf{Input:} failure $F$, success set $\mathcal{S}$ for same $(\textit{domain},\textit{task\_id})$, domain $d$
\STATE \textbf{Output:} $(x,y^+,y^-)$ or $\varnothing$
\STATE \textbf{Stage 1 — State-based divergence}
\STATE $\sigma_F \leftarrow \operatorname{ExtractState}(F)$
\STATE $\textit{best} \leftarrow \varnothing$; $\textit{best\_score} \leftarrow 0$
\FORALL{$s \in \mathcal{S}$}
  \FORALL{assistant turn $j$ in $s$}
    \STATE $\sigma_s \leftarrow \operatorname{ExtractState}(s,j)$
    \IF{$\operatorname{Compatible}(\sigma_F, \sigma_s)$}
      \STATE $\textit{score} \leftarrow \operatorname{Jaccard}(\sigma_F.\textit{fields}, \sigma_s.\textit{fields})$
      \IF{$\textit{score} > \textit{best\_score}$}
        \STATE $\textit{best\_score} \leftarrow \textit{score}$; $\textit{best} \leftarrow (s,j)$
      \ENDIF
    \ENDIF
  \ENDFOR
\ENDFOR
\IF{$\textit{best} \neq \varnothing \;\text{and}\; \operatorname{ActionAt}(F) \neq \operatorname{ActionAt}(\textit{best}.s, \textit{best}.j)$}
  \STATE $x \leftarrow \operatorname{System}(d)\oplus\operatorname{TG}(d)\oplus \textit{best}.s_{<j}$
  \STATE $y^+ \leftarrow \operatorname{ActionAt}(\textit{best}.s,j)$; $y^- \leftarrow \operatorname{ActionAt}(F)$
  \IF{$\operatorname{Tok}(x)+\operatorname{Tok}(y^+)>8192$} \RETURN $\varnothing$ \ENDIF
  \RETURN $(x,y^+,y^-)$
\ENDIF
\STATE \textbf{Stage 2 — Prefix-based divergence} \COMMENT{Stage~1 matched no compatible state}
\STATE Pick any $s \in \mathcal{S}$ with $R(s) > R(F)$
\STATE $c \leftarrow \operatorname{LCP}_{50}(s,F)$
\IF{$c=0$} \RETURN $\varnothing$ \ENDIF
\IF{$\operatorname{RoleAt}(s,c)=\textsc{User}$} \STATE $c\leftarrow c+1$ \ENDIF
\STATE $k \leftarrow \operatorname{ToolCount}(s_{<c})$
\IF{$\neg\operatorname{AllMatch}(s_{<c}) \;\text{or}\; \operatorname{Match}(F_k)$} \RETURN $\varnothing$ \ENDIF
\STATE $y^+ \leftarrow \operatorname{NextAsst}(s,c)$; $y^- \leftarrow \operatorname{NextAsst}(F,c)$
\IF{$y^+=\varnothing \;\text{or}\; y^-=\varnothing$} \RETURN $\varnothing$ \ENDIF
\IF{$\operatorname{SameAction}(y^+,y^-) \;\text{or}\; \operatorname{Sim}(y^+,y^-)>0.80$} \RETURN $\varnothing$ \ENDIF
\STATE $x \leftarrow \operatorname{System}(d)\oplus\operatorname{TG}(d)\oplus s_{<c}$
\IF{$\operatorname{Tok}(x)+\operatorname{Tok}(y^+)>8192$} \RETURN $\varnothing$ \ENDIF
\RETURN $(x,y^+,y^-)$
\end{algorithmic}
\end{algorithm}

\textbf{Prefix-based divergence.} When state matching fails but success trajectories exist for the task, the framework falls back to Stage~2: it aligns the two trajectories message-by-message via $\operatorname{LCP}_{50}$ to locate the divergence point. $\operatorname{LCP}_{50}$ scans aligned messages comparing each message's role and first 50 content characters to find the shared prefix $c$. The annotation-based correctness filter ($\operatorname{AllMatch}$ and $\operatorname{Match}$) discards pairs where any chosen-path action before divergence is annotated as incorrect, or where the rejected divergence action is annotated as correct. $\operatorname{NextAsst}$ extracts the diverging assistant messages as $y^+$ and $y^-$; $\operatorname{SameAction}$ and $\operatorname{Sim}$ (the \texttt{SequenceMatcher} ratio) filter out identical actions and near-duplicates. Both paths produce pairs in the same format, and pairs are deduplicated per task (at most five per task), totaling 161 pairs with a train-validation split by task ID. If no success trajectory exists for the task, the failure case is discarded.

\textbf{TG context consistency.} In v1, DPO pairs were constructed with a bare system prompt while inference used the full TG-injected prompt; this mismatch coincided with degraded online reward. In v2, the constructor loads the same weighted ToolGraph used at inference and injects its planning prompt into the training data:

\begin{quote}\small\ttfamily
\raggedright system\_content = f"\{tools\_text\}"\textbackslash n\textbackslash n tg\_get\_planning\_prompt()
\end{quote}

This supplies the same structural context during training and inference, avoiding the prompt mismatch observed in v1.

\textbf{Training and deployment.} Preference pairs are trained using 4-bit NF4 QLoRA with double quantization and bfloat16 compute. Rank-16 LoRA adapters use $\alpha=32$ and dropout 0.05, targeting the Qwen attention projections (\texttt{q/k/v/o}) and MLP projections (\texttt{gate/up/down}). Training uses a per-device batch size of 1, gradient accumulation of 8, a maximum sequence length of 4096, and a 0.1 warm-up ratio. We optimize the standard DPO loss:

\[
\Delta_\theta(x,y)=
\log \pi_\theta(y \mid x) - \log \pi_{\text{ref}}(y \mid x),
\]
\[
\mathcal{L}_{\text{DPO}}
= -\mathbb{E}_{(x,y_w,y_l)\sim\mathcal{D}}
\left[
\log \sigma\left(
\beta(\Delta_\theta(x,y_w)-\Delta_\theta(x,y_l))
\right)
\right].
\]

where $\beta$ controls deviation from the reference model. The 16 evaluated configurations span four $\beta$ values (0.05, 0.10, 0.15, and 0.20), three learning rates ($2$, $5$, and $8\times10^{-6}$), and four training budgets (60, 100, 150, and 200 steps); they do not form a full factorial grid. The deployed R2\#4 checkpoint uses $\beta=0.20$, learning rate $5\times10^{-6}$, and 100 training steps. Among the evaluated configurations, it combines positive chosen reward with the highest observed end-to-end weighted reward. Table~\ref{tab:dpo-summary} shows representative configurations.

\begin{figure*}[t]
  \centering
  \includegraphics[width=0.78\textwidth]{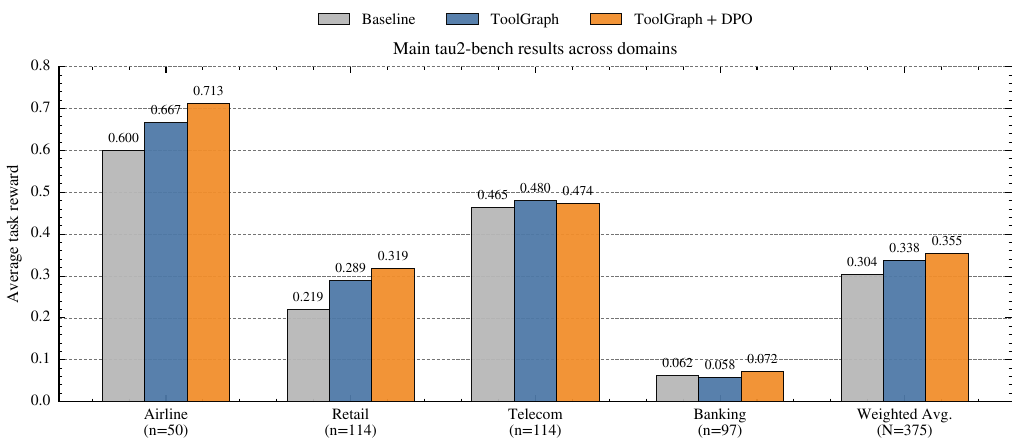}
  \Description{Grouped bar chart comparing Baseline, ToolGraph, and ToolGraph plus DPO across airline, retail, telecom, banking, and weighted average rewards.}
  \caption{Main tau2-bench results across four domains. Parenthetical values indicate task counts. ToolGraph improves the weighted average over the baseline, and DPO R2\#4 further improves airline, retail, banking, and the weighted average.}
  \label{fig:main-results}
\end{figure*}

\begin{table}[ht]
\caption{Selected DPO training configurations from the 16 evaluated settings. In these runs, negative chosen reward is associated with catastrophic forgetting despite high margins.}
\label{tab:dpo-summary}
\centering
\footnotesize
\begin{tabular}{lcccccc}
\toprule
Config & $\beta$ & LR & Steps & Chosen & Margin & Acc. \\
\midrule
R2\#4 (deployed) & 0.20 & $5{\times}10^{-6}$ & 100 & $+$0.03 & 2.83 & 0.93 \\
R2\#5 (overfit)  & 0.20 & $5{\times}10^{-6}$ & 150 & $-$0.28 & 3.93 & 0.91 \\
R2\#6 (catastrophic) & 0.15 & $8{\times}10^{-6}$ & 150 & $-$1.16 & 5.05 & 0.94 \\
\bottomrule
\end{tabular}
\end{table}

The trained LoRA adapter is merged into the base model using PEFT's \texttt{merge\_and\_unload()}, with Multi-Token Prediction heads copied from the base model to avoid vLLM loading errors. The checkpoint comparison is analyzed in Section~\ref{sec:dpo-analysis}.


\section{Experiments}

\subsection{Protocol}

We evaluate all methods on the same four tau2-bench domains: airline (50 tasks), telecom (114 tasks), retail (114 tasks), and banking (97 tasks), totaling 375 tasks. The base model is Qwen/Qwen3.5-9B deployed through vLLM on HPC2 A800 GPUs. Inference uses temperature 0.7, top-$p$ 0.8, a 2048-token generation limit, a 50-step interaction budget, up to three candidate generations per decision, and at most four concurrent simulations; failed model requests are retried once. We compare three configurations: (1) baseline with no ToolGraph intervention, (2) ToolGraph (TG) with curated edge weights, pre-act gate, loop check, and Critical Rules, and (3) TG + DPO R2\#4 adapter. Edge weights and DPO pairs are constructed from the agent's own prior rollouts on this benchmark task set, so the results demonstrate within-benchmark self-improvement rather than zero-shot generalization to unseen tasks. TG-based methods are evaluated over three rounds; the baseline is one available run.

\subsection{Main Results}

Figure~\ref{fig:main-results} shows the main results. ToolGraph raises the weighted average reward from 0.304 to 0.338 (+0.034), and DPO R2\#4 further improves it to 0.355 (+0.017). Gains concentrate in airline (0.600 $\rightarrow$ 0.667 $\rightarrow$ 0.713) and retail (0.219 $\rightarrow$ 0.289 $\rightarrow$ 0.319). Telecom shows little change in our runs (0.465--0.480), and banking remains low (0.058--0.072).

Comparisons of per-task mean rewards show complementary changes from TG and DPO. In airline, TG improves 11 tasks and regresses on 9 relative to the baseline; DPO then improves 12 and regresses on 7 relative to TG. In retail, the corresponding counts are 36 vs.\ 19 for TG and 28 vs.\ 18 for DPO. This pattern is consistent with TG addressing structural failures and DPO refining strategy on the margin.

\subsection{Error Analysis}

\begin{table}[t]
\caption{Selected error rates by method and domain. Values are percentages of all tasks in the corresponding runs with each error label, computed from the run summaries; labels are non-exclusive.}
\label{tab:error-rates}
\centering
\scriptsize
\setlength{\tabcolsep}{3pt}
\begin{tabular}{llrrrr}
\toprule
Method & Domain & Duplicate & Low reward & Max steps & Missing args \\
\midrule
Baseline & Airline & 78.0 & 40.0 & 12.0 & 4.0 \\
Baseline & Retail  & 57.9 & 48.2 & 7.0  & 0.9 \\
Baseline & Telecom & 98.2 & 53.5 & 52.6 & 2.6 \\
Baseline & Banking & 87.6 & 92.8 & 41.2 & 26.8 \\
\midrule
TG & Airline & 72.0 & 28.7 & 2.0 & 2.0 \\
TG & Retail  & 54.7 & 35.7 & 1.8 & 0.0 \\
TG & Telecom & 98.2 & 51.5 & 48.8 & 2.0 \\
TG & Banking & 65.3 & 61.2 & 19.9 & 24.4 \\
\midrule
DPO R2\#4 & Airline & 74.0 & 26.7 & 2.0 & 6.0 \\
DPO R2\#4 & Retail  & 54.7 & 33.0 & 0.3 & 0.3 \\
DPO R2\#4 & Telecom & 99.4 & 52.3 & 50.3 & 1.8 \\
DPO R2\#4 & Banking & 68.7 & 63.6 & 21.3 & 25.8 \\
\bottomrule
\end{tabular}
\end{table}

Table~\ref{tab:error-rates} shows that ToolGraph reduces \texttt{MAX\_STEPS} terminations in airline (12.0\% to 2.0\%) and retail (7.0\% to 1.8\%). Low-reward errors also fall from 40.0\% to 28.7\% in airline and from 48.2\% to 35.7\% in retail; DPO further lowers them to 26.7\% and 33.0\%. These changes are consistent with the complete ToolGraph controller: phase hints provide a structural prior, the pre-act gate checks write prerequisites, and weighted transitions guide beam scoring toward empirically successful successors. Because these mechanisms are evaluated jointly, the results support the full configuration but do not identify each component's independent effect.

Telecom differs: its full-run \texttt{MAX\_STEPS} rate remains high (52.6\% baseline, 48.8\% TG, 50.3\% DPO), making premature step-budget exhaustion its dominant observed failure mode. Banking improves on several error labels under TG, but low-reward and missing-argument rates remain substantially higher than in airline and retail.

\subsection{Domain-specific Diagnostics}

Reward computation differs across domains: airline and retail are entirely DB-state driven, banking is primarily DB-driven (88\%), and telecom uses environment assertions for all tasks. Consequently, DB match is not a valid success proxy for telecom. In the TG \texttt{USER\_STOP} subset, 114 of 170 telecom trajectories (67\%) achieve reward $\geq 1.0$ despite DB mismatch; 94\% of DB-mismatched trajectories still pass the environment assertion.

Telecom's bottleneck occurs before action execution. Nearly all full-run \texttt{MAX\_STEPS} cases have \texttt{action\_total=0}, whereas trajectories that reach \texttt{USER\_STOP} show 95--96\% action all-OK and 97--99\% environment-assertion pass rates under TG and DPO. The agent often exhausts its 50-step budget querying multiple line identifiers instead of deciding when to act. ToolGraph's loop check reduces this behavior only modestly, and divergence-point DPO provides no direct signal for trajectories that never reach an action. Demonstrations of the gather-to-action transition may therefore be more appropriate for this failure mode.

Banking remains difficult for the evaluated 9B model (0.062 baseline, 0.058 TG, 0.072 DPO). Its TG diagnostic subset has 15\% DB pass and 13\% action all-OK, the lowest among the four domains. The domain combines knowledge-base search, unlockable tool flows, financial-policy reasoning, and multi-step verification; even frontier models achieve only approximately 25\% Pass\textsuperscript{1}~\citep{tauknowledge2026}. The present single-model evaluation cannot separate model-capability constraints from method-specific limitations.

\subsection{DPO Training Analysis}
\label{sec:dpo-analysis}

We evaluated 16 DPO configurations spanning $\beta$, learning rate, and training steps on the full tau2-bench. Table~\ref{tab:dpo-summary} shows three representative settings. R2\#4 improves airline, retail, and banking; R2\#5 overfits, with banking collapsing to 0.01; and R2\#6 causes catastrophic forgetting in airline and retail.

The DPO reward clarifies this behavior:
\[
r_\theta(y_w \mid x)
= \log \pi_\theta(y_w \mid x)
- \log \pi_{\text{ref}}(y_w \mid x).
\]
A negative chosen reward means the trained model assigns lower probability to a correct response than the frozen reference model. High margin can therefore coexist with suppression of previously correct behavior. Across these configurations, chosen reward positivity is more informative than margin or accuracy for checkpoint selection.

Prompt alignment is the first mechanism behind the deployed checkpoint's improvement. The v1 adapter, trained without the ToolGraph prompt but deployed with it, reduced online reward from 0.349 to 0.296. The v2 pairs retain the deployment-time ToolGraph context, avoiding this train--deployment mismatch.

Reward-info filtering is the second mechanism. It removes candidates when available annotations reveal a chosen-path error before divergence or mark the rejected divergence action as correct. The filter removes 26\%, 57\%, 14\%, and 12\% of candidate pairs in airline, retail, telecom, and banking, respectively. This filter uses the benchmark's action-level correctness annotations to isolate causal preference signals: pairs are excluded when the annotation reveals that the chosen-path error occurred before divergence, or that the rejected divergence action is actually correct.

The sub-scores suggest different forms of improvement across domains. In retail, DPO raises action all-OK from 41\% to 46\% and DB pass from 46\% to 49\%. In airline, action all-OK remains 46\% while reward rises from 0.667 to 0.713, suggesting a sequence-level strategy gain rather than a higher per-call match rate.


\FloatBarrier
\section{Conclusion}

This paper shows that a local 9B agent can improve its own tool-use reliability on tau2-bench by reusing benchmark execution experience as both structured inference-time guidance and prompt-aligned preference supervision. The combined ToolGraph and DPO pipeline raises the weighted average reward from 0.304 to 0.355, with the DPO gain concentrated in airline and retail. Two insights emerge beyond the headline numbers.

First, different domains exhibit different bottlenecks in the available fine-grained diagnostics. Telecom's DB pass rate (29\%) is misleading in the \texttt{USER\_STOP} diagnostic subset: action and environment checks pass 95--97\% of the time once the agent reaches the action phase, while 49--53\% of full-run trajectories terminate at \texttt{MAX\_STEPS}. Second, chosen reward positivity is the most useful DPO health diagnostic observed across our 16 evaluated configurations: it distinguishes the deployed checkpoint from configurations exhibiting catastrophic forgetting despite high accuracy or margin.

Banking remains difficult for the evaluated 9B model. Future directions include a targeted SFT dataset teaching ``when to stop gathering'' for telecom, iterating the self-evolution loop, and evaluating larger base models to separate orchestration effects from model capability.

\subsection{Limitations}

All trajectory experience and evaluation come from the same tau2-bench task set, so the results demonstrate within-benchmark self-improvement rather than generalization to unseen tasks or domains. ToolGraph's mechanisms are evaluated jointly; without component-wise ablation, the results support the combined configuration rather than the independent contribution of each mechanism. Phase classification, Critical Rules, and scoring parameters are manually designed rather than learned automatically. Only Qwen/Qwen3.5-9B is evaluated, with a single baseline run; without multiple runs or model scales, we cannot report error bars.

\bibliographystyle{ACM-Reference-Format}
\bibliography{references}

\end{document}